\documentclass[runningheads]{styles/llncs}
\usepackage{graphicx}
\usepackage{comment}
\usepackage{amsmath,amssymb} 
\usepackage{color}
\usepackage{adjustbox}
\usepackage{subcaption}
\usepackage{amssymb}
\usepackage{pifont}
\usepackage{multirow, boldline}
\usepackage{ctable}
\usepackage{xcolor}
\usepackage[bottom]{footmisc}

\usepackage[pagebackref=true,breaklinks=true,letterpaper=true,colorlinks,bookmarks=false]{hyperref}
\usepackage{breakcites}
\hypersetup{
     colorlinks   = true,
     citecolor    = green
}
\hypersetup{linkcolor=red}

\newcommand{\etal}{\textit{et al}.}
\newcommand{\ie}{\textit{i.e.} }
\newcommand{\eg}{\textit{e.g.} }
\newcommand{\ours}{DMV}
\newcommand{\ourss}{DMV }

\definecolor{darkgreen}{rgb}{0.0, 0.6, 0.2}

\newcommand{\Tref}[1]{Table~\ref{#1}}

\newcommand{\Eref}[1]{Equation~(\ref{#1})}
\newcommand{\fref}[1]{Fig.~\ref{#1}}
\newcommand{\Fref}[1]{Figure~\ref{#1}}

\newcommand{\Sref}[1]{Section~\ref{#1}}

\begin{document}
\pagestyle{headings}
\mainmatter
\def\ECCVSubNumber{3714}  

\def\JL#1{{\color{red}JL: \it #1}}

\title{DMV: Visual Object Tracking via Part-level Dense Memory and Voting-based Retrieval}

\titlerunning{DMV}
%
\author{Gunhee Nam\inst{1} \and
Seoung Wug Oh\inst{1} \and
Joon-Young Lee\inst{2} \and
Seon Joo Kim\inst{1,3}}
\authorrunning{Nam et al.}
%
\institute{Yonsei University, Seoul, South Korea \and
Adobe Research, San Jose CA, USA \and
Facebook, San Jose CA, USA}
\maketitle

\begin{abstract}

We propose a novel memory-based tracker via part-level dense memory and voting-based retrieval, called DMV.
Since deep learning techniques have been introduced to the tracking field, Siamese trackers have attracted many researchers due to the balance between speed and accuracy.
However, most of them are based on a single template matching, which limits the performance as it restricts the accessible information to the initial target features.
In this paper, we relieve this limitation by maintaining an external memory that saves the tracking record.
Part-level retrieval from the memory also liberates the information from the template and allows our tracker to better handle the challenges such as appearance changes and occlusions. 
By updating the memory during tracking, the representative power for the target object can be enhanced without online learning.
We also propose a novel voting mechanism for the memory reading to filter out unreliable information in the memory.
We comprehensively evaluate our tracker on OTB-100~\cite{otb2015}, TrackingNet~\cite{trackingnet}, GOT-10k~\cite{got10k}, LaSOT~\cite{lasot}, and UAV123~\cite{mueller2016uav}, which show that our method yields comparable results to the state-of-the-art methods.

\end{abstract}
\section{Introduction}

Visual object tracking is a task of detecting and tracking the target object through a whole video sequence relying on visual information of the target given at the beginning of the sequence.
Recently, the table of this long standing problem of the computer vision field has been turned with deep learning techniques ~\cite{krizhevsky2012alexnet,he2016resnet,huang2017densenet,faster_rcnn,liu2019cpnet}. 
However, one of the challenges in training convolutional networks for tracking, different from recognition tasks, is that the target object is specified at the test time.
Online learning is a common solution to deal with this challenge.
The network parameters are fine-tuned at the test time ~\cite{mdnet,real-time_mdnet,danelljan2019atom} or discriminative filters are updated depending on its estimation~\cite{bolme2010corr_filter,eco,ccot}.
The networks learning online have to be light-weighted for real-time performance, because the back-propagation for the parameter update takes a considerable amount of time. 
Therefore, online learning is an obstacle to harnessing the deep representational power of convolutional networks.
In addition, online learning is often sensitive to its hyper-parameters, which leads to unstable performance.

Meanwhile, Siamese trackers showed that deep networks trained offline can also run in real-time with competitive performance to online trackers~\cite{siamfc,siamrpn,siamrpn++}.
These approaches interpret the tracking task as a matching problem between the template and the search images in an embedding space by the cross-correlation.
Yet, the visual variations of the target object are hard to be recognized by the Siamese trackers because the template contains only the target information in the initial frame.
For this reason, recent deep tracking methods consider adopting memory mechanism.
The use of memory for tracking has advantages in that various appearances of the target object can be memorized without vanishing through time. 
However, previous memory-based trackers are built on top of the Siamese framework limiting its full potential~\cite{yang2018memtrack,yang2019memdtc,lee2018mmlt,shi2019dawn,baik2019p2fnet}. 
These trackers usually use memory information to adapt the initial template of Siamese trackers to the current state.
The memory information is loaded without considering the changes in the target's geometric properties (\eg the location of each part). 
Thus they cannot make use of distant frames, when the target's appearance has changed a lot, for adapting the template.
Therefore, previous memory trackers ironically use only adjacent frames for memory, ceiling the core effect of the memory usage.

\begin{figure}[t]
\begin{center}
\begin{subfigure}[b]{0.49\linewidth}
\centering
  \includegraphics[width=0.95\linewidth]{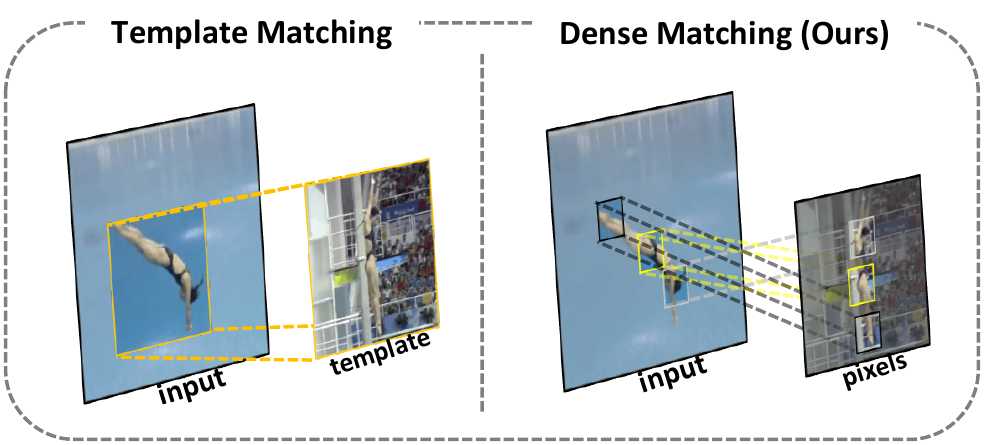}
\end{subfigure}
\begin{subfigure}[b]{0.49\linewidth}
\centering
  \includegraphics[width=0.95\linewidth]{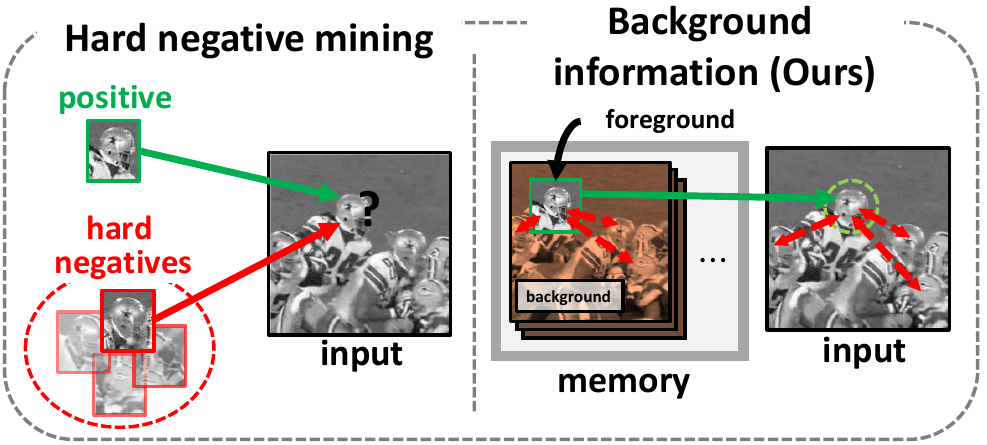}
\end{subfigure}
\end{center}
\vspace{-5mm}
\caption{Comparison of our method with existing methods. (\textit{Left}) Siamese-based trackers rely on template matching, while our method has an external memory and employs the attention mechanism for dense matching.
This enables us to use intermediate information effectively and be robust to deformation and appearance changes. (\textit{Right}) We store both the foreground and the background information in a unified memory, which helps our tracker to differentiate the target using semantic information.}
\label{fig:intro}
\end{figure}

To address the aforementioned issues, we propose a novel attention-based tracker that takes advantage of the memory of estimated intermediate frames.
We use external memory to store mid-level features of history frames which enables us to represent part-level information. We employ attention mechanisms for dense feature matching which enables us to fully exploit the memory information.
Our memory representation and feature matching resolve the structural constraint of the previous template-based matching that compares the whole objects in the template.
The difference between our dense matching and conventional template matching is illustrated in \Fref{fig:intro}.
Our attention-based matching is performed on part-level, and it makes our model more robust to the changes in the target's geometric properties. 
Meanwhile, the bounding box label is too coarse to derive the fine annotation for the dense target information, therefore handling noisy annotations is crucial for the fine-scale matching.
In order to handle the noise issue, we adopt a voting system to our memory retrieval module. Through our voting system, the most similar matching candidates are selected and compared with each other to confirm the reliability.
With the voting system, our networks learn to differentiate unreliable information and selectively retrieve reliable information from all the available sources.
In addition, by encoding the uncertainty of estimation when writing into the memory, the effect of noisy annotations is further decreased.

Hard negative mining is also considered as one of the most important tricks for visual tracking~\cite{zhu2018dasiam,mdnet,danelljan2019atom}.
In many cases, templates for hard negative candidates are sampled and maintained by heuristics to alleviate drifting.
In our framework, on the other hand, explicit hard negative mining is not required as the similar functionality comes at no cost.
By writing the whole image with label information on the memory, the networks become aware of negative samples through end-to-end learning.
As illustrated in \Fref{fig:intro}, semantic information of both the target and the background helps our tracker to differentiate the target from the hard negative samples.
Furthermore, this system allows efficient unified memory for both the foreground and the background, while other memory trackers~\cite{yang2019memdtc,shi2019dawn} have separate memory slots for each of them.

We tested our tracker on numerous recent tracking benchmarks including
OTB-100~\cite{otb2015}, TrackingNet~\cite{trackingnet}, GOT-10k~\cite{got10k}, LaSOT~\cite{lasot}, and UAV123~\cite{mueller2016uav}.
Our tracker achieves top performance with other state-of-the-art trackers, while keeping real-time performance ($>$40 FPS).
In addition, we compare different memory architectures and experimentally validate the efficacy of our method.

Contribution of this paper is summarized as follows:
\begin{itemize}
    \item We propose a novel attention-based tracker that takes advantage of a memory with multiple estimated frames through dense feature matching.
    \item In order to handle noisy memory information, we present a memory retrieval module to load the most reliable information based on the voting system.
    \item We propose a natural way to be aware of hard negative samples by retrieving background information densely without heuristic hard negative mining. 
    \item We perform comprehensive experiments and achieve state-of-the-art results on various tracking benchmarks.
\end{itemize}

\section{Related Work}
Visual object tracking has a long history
with various branches of the problem in terms of their properties~\cite{black1996eigentracking,comaniciu2000meanshift,avidan2004svt,babenko2009mil,kalal2010pn}.
As deep learning has shown its potential in other areas in computer vision, it has also become the dominant tool for visual tracking. 
Siamese networks and online learning methods have especially shown to be effective for visual tracking.
In this section, we give a brief review on these deep trackers.
Additionally, we separately review memory-based trackers, where estimated intermediate frames are written on the memory to boost the discriminative power of trackers without leaking information.
We refer to the survey papers~\cite{yilmaz2006survey,wu2013online_survey,li2018deep} for further discussion on visual object tracking.

\subsection{Online Learning Trackers}
Online learning is one of the traditional methods for visual object tracking~\cite{kalal2009online,zhang2013sparse_rep,jang2015error}.
Because trackers need to track an arbitrary object only by its appearance given at the start, online learning is a natural way to fit the model to the object as time passes.
In the deep learning era, this method became one of the main strategies to compensate for the limitation of the offline learning~\cite{mdnet,real-time_mdnet,song2018vital,danelljan2019atom}.
Nam~\etal~\cite{mdnet} treated the tracking problem as a classification problem by an online learning of simple discriminator.
Randomly sampled boxes near the location of a target object are independently classified by the discriminator to determine whether it is a target or a background.
As one of the problems in online learning is the low speed due to the backtracking, Jung~\etal~\cite{real-time_mdnet} improved the classification process by employing RoI pooling to reduce the repeated operations of the previous work.
Song~\etal~\cite{song2018vital} imported adversarial learning into tracking by generating a weight mask for better classification.
IoU predictor in ATOM~\cite{danelljan2019atom} improved the performance of the tracking by locating the target more accurately.
Unlike the online trackers, our method tracks the target without online learning.
Tracking by offline learning only makes our model faster and more stable without the changes in parameters during the test time. 

\subsection{Siamese Trackers}
Siamese trackers adopt Siamese networks into their architectures as a one-shot learner.
Since the purpose of visual object tracking can be regarded as the problem of detecting an object which has not been trained on before, the property of the Siamese networks matches well with the goal of tracking~\cite{koch2015siamese}.
They show high speed and accuracy despite its simple architecture~\cite{siamfc,siamrpn,siamrpn++,zhu2018dasiam}.
Li~\etal~\cite{siamfc} first demonstrated that Siamese networks are effective for visual tracking.
By computing the cross-correlation between the two features from an exemplar and an instance frame, it predicts a response map for the target object.
However, the work~\cite{siamfc} has disadvantages in that it only estimates the center of the target object without the changes in the scale and the ratio of the bounding boxes. 
To overcome this problem, Li~\etal~\cite{siamrpn} proposed a region proposal framework for the Siamese networks. 
Following this work, Li~\etal~\cite{siamrpn++} improved the performance of the Siamese tracker again by applying deeper neural networks and taking advantage of multiple features from various levels of the backbone networks.
The discriminative capacity of SiamRPN~\cite{siamrpn} was further enhanced in DaSiamRPN~\cite{zhu2018dasiam} by training the model to be aware of the negative samples as well as the positive samples.
However, because the aforementioned trackers utilize only the initial frame as the template, they have difficulties in dealing with the appearance changes of the target.

\subsection{Memory-based Trackers}
To recognize intermediate changes of the target, various methods have attempted to make use of the intermediate estimations.
Traditionally, exploiting the predicted intermediate frames to bootstrap trackers has been consistently considered.
Babenko~\etal~\cite{babenko2009mil} proposed a method to utilize multiple instance patches as the training set of the tracker, in order to avoid problems caused by incorrectly labeled instances.
Kalal~\etal~\cite{kalal2010pn} also addressed the same issue through the P-N learning, in which both of the positive and the negative constraints guide the identification of exemplars.

Many deep learning approaches also tried to utilize estimated information.
Zhu~\etal~\cite{zhu2018flowtrack} aligned templates from multiple frames according to their estimated optical flow to generate the adaptive template.
Lee~\etal~\cite{lee2018mmlt} proposed memory trackers for long term tracking by interpolating the multiple estimated templates stored in a short term and a long term memory according to the number of successive frames with reliable results. 
Dynamic memory networks for object tracking controlled by an LSTM controller to retrieve reliable information from the memory were presented in \cite{yang2018memtrack,baik2019p2fnet}.
The memory tracker was further extended in \cite{yang2019memdtc,shi2019dawn} by a revised loss function, an intuitive attention module, and the background information.
These memory-based approaches rely on Siamese-based template matching, where the spatial constraints from the template limit the full utilization of the memory information. On the other hand, we densely encode mid-level features for the memory information and employ attention-based matching, which enables us to perform part-level matching and leads to high performance.

\subsection{Memory Networks}
Memory networks is a neural network coupled with external memory where information can be stored and read whenever necessary~\cite{sukhbaatar2015endtoend,kumar2016dynamicmemory}.
Recently, it is widely adopted for various vision tasks such as video inpainting~\cite{oh2019onion}, movie story understanding~\cite{na2017read}, and video object segmentation~\cite{memory_seg}.
Among them, our method is largely inspired by the space-time memory networks (STM) in~\cite{memory_seg}.
However, STM is not directly applicable to visual tracking due to an important difference between video object segmentation and tracking.
While pixel-accurate fine annotation is given in video segmentation task, the bounding box annotation in tracking is coarse and inaccurate, thus bounding boxes always contain background pixels. 
To address the noisy annotation issue, we redesign the memory retrieval module, which is the heart of memory networks.
Specifically, we propose a novel voting mechanism for selecting reliable information with uncertain annotation.

\section{Method}
\begin{figure*}[t]
\begin{center}

\includegraphics[width=1\linewidth]{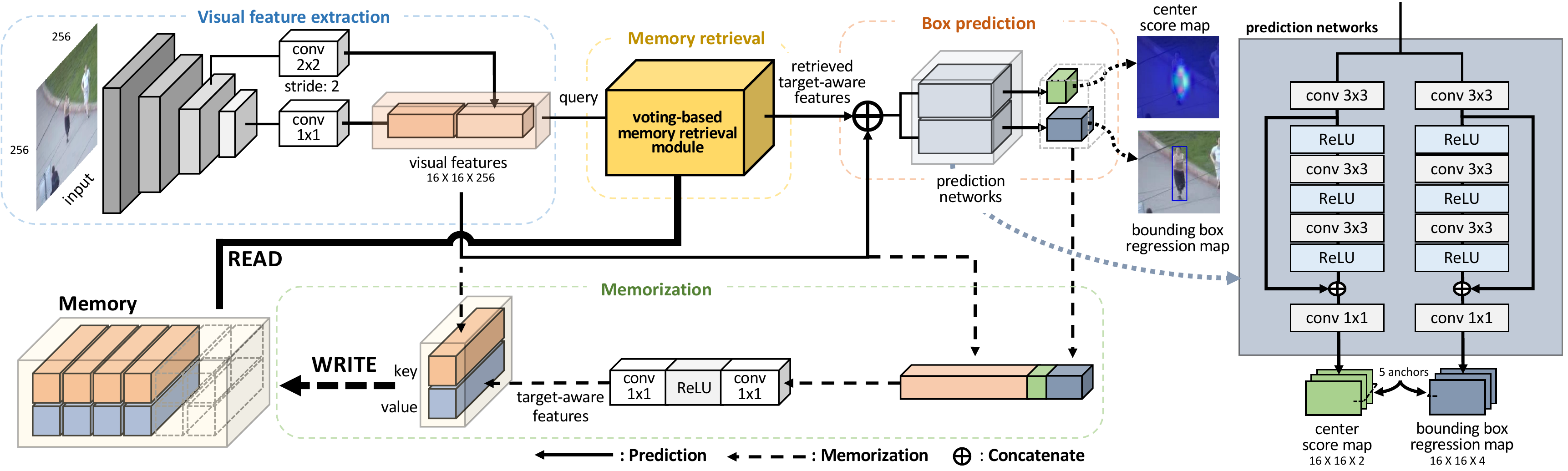}
\end{center}
  \vspace{-0.4cm}
   \caption{The overview of our \ourss tracker.}
  \vspace{-0.6cm}
\label{fig:overview}
\end{figure*}
Our method processes a video frame-by-frame in a time sequence. 
We consider previous video frames with bounding box annotations as memory, and the current frame as an input. 
\Fref{fig:overview} shows the overview of our model. 

Our tracker is initialized by writing the first frame with a given box annotation onto the memory.
At each memory slot, we store a pair of \textit{key} and \textit{value}, which are deep features with different roles.
\textit{Key} is a set of visual features extracted from deep backbone network, and \textit{value} is a set of target-aware features, integrating visual and localization information (e.g. bounding boxes).

For visual tracking, the visual features of the current frame, called \textit{query}, are first extracted by the backbone networks.
Then, the \textit{query} and every \textit{key} in the memory are densely matched to retrieve \textit{values} through our voting-based memory retrieval module. 
Finally, the visual features of the current frame (\textit{query}) and the retrieved target-aware features (\textit{value}) are used together to predict the location of the target object through the box prediction head.
During tracking, the memory is updated by adding the prediction results. 
Here, we explain the details of each component.

\subsection{Main components} \label{subs:write}

\noindent\textbf{Visual feature extraction.} 
The visual features, query and key, are extracted through the shared backbone networks. The `query' represents the visual features extracted from the input image. The `key' represents the visual features extracted from the previous frames and stored in the memory.
By using the shared backbone networks, query and key are encoded into the same embedding space, enabling effective matching at the retrieval module.
For the backbone, we use either \texttt{resnet18} or \texttt{resnet50} depending on the computational budget~\cite{he2016resnet}.
Specifically, we feed the features from \texttt{layer3} and \texttt{layer4} to a bottleneck convolution layer with and without stride in order to match the spatial resolution of two features, then concatenate them to get the query or key features.

\noindent\textbf{Memory retrieval.}
The visual features from two sources (`query' from the input and `key' in the memory) are densely compared to retrieve target-aware features (`value' in the memory) through the memory retrieval module. 
The details of our voting-based dense memory retrieval module is explained in~\Sref{subs:module}.

\noindent\textbf{Box prediction.}
To predict the bounding box, the visual features of the input (query) and the retrieved target-aware features from the memory (value) are used together.
In specific, the query and the retrieved value are concatenated and go through the prediction networks. 
Similar to~\cite{siamrpn}, it consists of two separated branches; one for estimating the center location of the target object and the other for the bounding box regression.
Each branch outputs the prediction for prefixed anchors.
The output of the center estimation branch is a binary probability map.
The bounding box regression branch estimates per-pixel four channel outputs for the anchor box displacements as defined in~\cite{faster_rcnn}.

\noindent\textbf{Memory update.}
After the box prediction, the memory is updated with a new estimate.
Each memory slot consists of a pair of `key' and `value' features.
The key is computed from shared backbone networks as explained. To encode the `value', we first concatenate the key feature map, the center score map, and the bounding box regression map, then feed that into two convolution layers with a ReLU activation.
To exploit background information, the information stored in the memory includes both the target object and the background. We use soft representation of the center score and box regression maps in order to encode the uncertainty of the estimate, instead of hard box annotation.
To minimize false positives, we set the score values lower than 0.5 to zero before encoding the value.
For the initial exemplar frame coming with the ground truth bounding box, the memory is formed by converting the hard box annotation into our soft representation (\ie one-hot classification score map with the exact anchor displacement).

\noindent\textbf{Relation to space-time memory network.}
Overall process of our framework resembles the video object segmentation framework proposed in~\cite{memory_seg}.
While our method is inspired by~\cite{memory_seg}, we made significant modification and improvement to realize the similar concept for the visual tracking task. 
The main difference is in the memory retrieval operation, which is the core of memory networks. 
Specifically, we employ a voting system among memory candidates to filter out noisy information coming from coarse bounding box annotations.
In practice, even accurate box naturally contains background and occluded pixels.
The proposed voting-based dense memory retrieval module is explained in \Sref{subs:module} and validated in \Sref{subs:baseline}.

\begin{figure}[t]
\begin{center}

\includegraphics[width=1\linewidth]{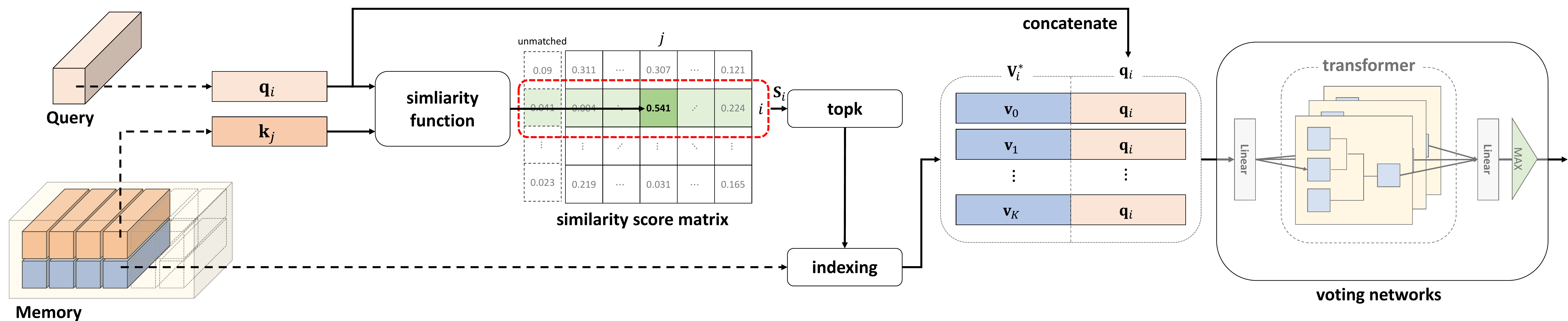}
\end{center}
\caption{The details of our voting-based memory retrieval module.}
\label{fig:module}
\end{figure}
\subsection{Voting-based Memory Retrieval Module} \label{subs:module}
The heart of our memory-based tracker is the voting-based memory retrieval module.
Our module for accessing the memory has three important properties:
\begin{itemize}
    \item \textbf{Dense query-key matching:} we match the query and key for all feature vector pairs, enabling the part-level matching in the global search range.
    \item \textbf{Learnable similarity metric:} this allows our tracker to learn the distance measure between feature vectors for the matching. 
    Once the metric is learned properly, our tracker has the online adaptation functionality with almost no additional cost, by simply augmenting the memory with new information.
    \item \textbf{Voting-based memory retrieval:} a number of candidate slots are selected from the memory, and the most reliable information is retrieved by comparing with each other. 
    By doing so, our model can learn to sort out noises while reading reliable information from the memory.
\end{itemize}

Our memory retrieval module takes the `query', and the `key' and `value' pairs from the memory as input, then returns the target-aware features relevant to the query.
The detailed illustration of the module is depicted in~\fref{fig:module}.
We explain the memory retrieval module step-by-step.

\noindent\textbf{Similarity score matrix.}
In our memory retrieval module, we first compute the similarity score matrix between all the pairs of the query and key feature vectors.
Each row of the similarity matrix $\mathbf{S}\in \mathbb{R}^{(HW)\times(THW+1)}$ is constructed as follow:
\begin{equation}
\label{eq:similarity}
    \mathbf{S}_{i} = [1, f(\mathbf{q}_i, \mathbf{k}_1), f(\mathbf{q}_i, \mathbf{k}_2), \dots, f(\mathbf{q}_i, \mathbf{k}_{THW})] / C,
\end{equation}
where $\mathbf{q} \in \mathbb{R}^{HW}$ and $\mathbf{k} \in \mathbb{R}^{THW}$ denote each feature vector of the query and key map, and $f(q, k)$ represents a similarity function.
We use the exponential dot-product similarity function, defined as $\exp(q \circ k)$ where $\circ$ denotes dot-product. The normalizing factor $C$ is $1+\sum_{\forall j}{\exp(\mathbf{q} \circ \mathbf{k}_j)}$.
We append one additional column to the similarity map (the first element of $\mathbf{S}_{i}$ in \Eref{eq:similarity}) to deal with the case when the query does not match with any keys.

\noindent\textbf{Memory candidates selection.}
After constructing the similarity matrix, memory candidates are selected according to the similarity scores.
The candidate value set $\mathbf{V}^*_i$ for each query location $i$ is determined by selecting the $K$ highest matching scores.
Namely, $\mathbf{V}^*_i = \{\mathbf{v}_{c} | c = \text{topk} ( \mathbf{S}_i ) \}$ where $\text{topk}(\cdot)$ outputs the indices with top $K$ scores for a given vector.
Different from the previous memory read operations~\cite{memory_seg,oh2019onion} that perform weighted-sum on the softmax output, our approach that selects top-$K$ candidates is advantageous in handling noises and hard negatives in the memory.
Our model can learn to reach a consensus among $K$ candidates through the following voting mechanism.

\noindent\textbf{Top-K Voting.}
After selecting candidates, the final target-aware features are retrieved by taking a vote to make a consensus among $K$ candidate values. 
To model a voting mechanism, the network needs to be able to process unstructured point data, being invariant to the order of input points. 
This leaves us only a few options~\cite{qi2017pointnet,vaswani2017attention} for the network design.
We take the Transformer~\cite{vaswani2017attention} as the backbone network for our voting networks. 

First, a set of candidate values $\mathbf{V}^*_i$ are concatenated with the original query $\mathbf{q}_i$, before fed into the voting network.  
In our voting networks, we wrap the multi-head Transformer model with two linear bottleneck layers to form a main body, to reduce computational overhead of the attention mechanism.
The main body is followed by max pooling to select the strongest responses.

Note that this is not the first trial for replacing the softmax function with the top-k operation in the attention block. 
However, we argue that the previous attempt~\cite{liu2019cpnet} is motivated by different insights for different tasks (\ie self-attention mechanism for action recognition).
Furthermore, the message exchange between candidates through our voting network is essential to read reliable information from the memory.
The effectiveness of our voting network is further discussed in \Sref{subs:baseline}.

\subsection{Loss} \label{subs:loss}
When training the networks, two separated loss functions are used, one for the bounding box center prediction ($L_c$) and the other for the bounding box regression ($L_b$).
As we adopt the anchor-based box estimation~\cite{faster_rcnn}, the losses are computed for each anchor in the anchor set $A$.   
Our final loss $L$ is the combination of those two losses with a balancing factor $\lambda$:
\begin{equation}
    L = \sum_{A}( L_c + \lambda L_b).
\end{equation}

\noindent\textbf{Box center prediction loss ($L_c$):}
The ground truth for the target object center is determined by the intersection over union (IoU) between the anchor box and the ground truth bounding box.
Empirically, imbalance between the numbers of the positive and the negative anchors causes difficulties to train the networks.
To balance between them, we use the same amount of anchors for each to compute the loss function by sampling the negative anchors with high prediction scores (\ie hard negatives).
A set of the positive center points $S_{pos}$ and a set of the negative ones $S_{neg}$ are selected as:
\begin{equation}
    S_{pos} = \{ i | y^c_i = 1 \},\quad
    S_{neg} = \{ i | y^c_i = 0,\text{ and } i\in\text{topk}(x^c_i)\},
\end{equation}
where $y^c_i$ denotes the ground truth label determined based on IoU, and $x^c_i$ is the center prediction of the networks. 
For the topk operation, $k$ is set to the number of the positive points ($|S_{pos}|$).

On the tracking task, the majority of the negative points are easy to distinguish (e.g. backgrounds pixels), and there are only few hard negatives (e.g. similar objects to the target).  
To further focus on the hard negatives and prevent the loss from being dominated by the majority of easy samples, we employ the focal loss~\cite{lin2017focal}.
Our box center prediction loss is defined as:
\begin{equation}
    L_c = \sum_{i \in S_{pos}}{(1-x^c_i)^2 \text{log}(x^c_i)}
        +\sum_{j \in S_{neg}}{(x^c_j)^2 \text{log}(1-x^c_j) }.
\end{equation}

\noindent\textbf{Bounding box regression loss ($L_b$):}
The bounding box regression loss is applied to fit the positive anchors to the ground truth box.
We take smooth $L_1$ as the distance measure for the regression~\cite{girshick2015fast_rcnn}. 
The loss is defined as:
\begin{equation}
    L_{b} = \sum_{i \in S_{pos}}{\text{smooth}_{L_1}(y^b_i - x^b_i)},
\end{equation}
where $y^b_i$ and $x^b_i$ are the ground truth and the network prediction for the bounding box regression, respectively.
Note that the bounding box regression loss is applied only for the positive anchor.

\subsection{Training} \label{subs:training}
To train our tracker, we sample $N$ frames with random skips from a training sequence.
Each training sample $T$ consists of $N$ tuples of a frame image, a ground-truth center score map and a bounding box regression map.
All of frames are rescaled according to the size of the target as in \cite{siamfc} to define a search region.
The first frame in $T$ is regarded as the initial frame, and the other frames are as the instance frames.
They are then forwarded to the tracker as described in \Sref{subs:write}.

During tracking, the further the frame is, the harder our model learns to discriminate the target object due to error accumulation and drifting.
To make our model robust to such problems, we give more weights on the loss for longer propagation by linearly scaling the significance according to the distance from the initial frame.
By doing so, our model can learn the better way to handle the uncertainty in the intermediate predictions. 

\noindent\textbf{Training details.}
Our model is optimized by stochastic gradient descent (SGD) with momentum~\cite{sutskever2013sgdmmt} of 0.9 and weight decay of 0.0005.
We use the initial learning rate of 0.001 decayed exponentially by 0.05 every 6.4k iterations.
Weights of the backbone networks is initialized with weights pretrained on ImageNet~\cite{russakovsky2015imagenet} and fine-tuned in the training time. 
We use the training set of GOT-10k~\cite{got10k} (9335 videos), LaSOT~\cite{lasot} (1280 videos), and TrackingNet~\cite{trackingnet} (30k videos) datasets.
Our model is trained for 32k iterations with a batch size of 16, taking about 2 days on a single NVIDIA RTX 2080 Ti.
The number of samples $N$ is gradually increased every 6.4k iterations from 2 until reaching 5 and the maximum skips between sampled frames are 100. 
Every sampled frame is randomly cropped containing the target object for its search space into size of 256 by 256 and augmented by stretching, blurring, graying and horizontal flipping randomly. 

\subsection{Inference} \label{subs:inference}
In the inference time, our tracker repeatedly performs the memory retrieval, box prediction, and memory update.
While having the information of intermediate predictions in the memory obviously enhances the performance of our tracker, storing all the information is not practically ideal because it increases the process time and the probability of memory pollution.
On the other hand, limited memory capacity can cause a lack of the diversity of the target appearance, limiting the potential of the memory.
For the best trade-off, we set the minimum time interval between memory frames to 30 frames and the maximum number of memory frames to 32 frames to increase the variance of the target appearance in the memory. 
If the memory reaches the full capacity, we drop the oldest memory frame out except for the initial frame that contains the most reliable information (\ie the ground truth).
To minimize the risk of memory pollution, we do not memorize the frame if the estimated center score is lower than a threshold of 0.7.
\section{Experiments}

We first compare different memory retrieval architectures and experimentally validate the efficacy of our module. Then, we conduct experiments on standard benchmarks to compare the performance of our method with state-of-the-arts.

%
\begin{table}[t]
\setlength{\tabcolsep}{3pt}
    
    \centering
    \begin{minipage}{.59\linewidth} 
      \centering
      \caption{Comparison with different memory retrieval modules.}
      \label{table:baseline}
      \begin{adjustbox}{width=1.0\linewidth}
        \begin{tabular}{l|c|c|c}
        \toprule
         & Softmax~\cite{memory_seg} & Top-k+MLP~\cite{liu2019cpnet} & Voting (Ours) \\
        
        \midrule
        
        OTB-100 (\%) & 53.5 & 61.2 & \textbf{69.9}\\
        UAV123 (\%) & 55.9  & 61.5 & \textbf{62.8}\\
        \bottomrule
        \end{tabular}
    \end{adjustbox}
    \end{minipage} \quad%
    \begin{minipage}{.36\linewidth}
      \centering
      \caption{Comparison of results for using memory.}
      \label{table:ablation}
      \begin{adjustbox}{width=1.0\linewidth}
        \begin{tabular}{c|cc}
        \toprule
        & No memory & Memory\\
        \midrule
         OTB-100 (\%) & 65.1 & \textbf{69.9} \\
         UAV123 (\%) & 61.5 & \textbf{62.8}\\
        
        \bottomrule
        \end{tabular}
    \end{adjustbox}
    \end{minipage}%
    \vspace{-0.4cm}
    
\end{table}

\subsection{Validation on our memory retrieval module}
\label{subs:baseline}

\noindent\textbf{Memory retrieval module.}
We design our memory retrieval module with two important considerations: part-level dense matching and learnable similarity metric. 
There are a few options that support such properties. 
We specifically consider the memory module in STM~\cite{memory_seg} and the attention block in CPNet~\cite{liu2019cpnet} as the most closely related alternative options.
To compare our module with the two architectures, we train our variant models after replacing the memory retrieval module and measure the AUC scores on the OTB-100~\cite{otb2015} and UAV123~\cite{mueller2016uav} datasets.
\Tref{table:baseline} shows the comparison result of different memory retrieval modules.
When we replace our memory module with the Softmax-based memory retrieval in STM~\cite{memory_seg}, it deteriorates the performance by 23.5\% and 11.0\% on both datasets, respectively.
We suspect that soft weights exacerbate the negative effects of noisy information. We hypothesize that the effects can be alleviated by selecting candidates rather than giving soft attention.
In the aspect of hard attention, CPNet~\cite{liu2019cpnet} performs a similar functionality with our module using the top-k operation. In \cite{liu2019cpnet}, each selected candidate is independently processed through the multi-layer perceptron (MLP).
We call this setting Top-k+MLP. 
This improves the performance to 61.2\% in the AUC score on the OTB-100 dataset with relative gains of 14.4\% compared with the Softmax-based module. 
This demonstrates selecting candidates is important for the tracking task where noisy bounding box annotations are given. However, this approach has a limitation in learning the interaction across candidates due to the independent processing of each candidate. 
On the other hand, our voting-based memory module facilitates the interaction across selected candidates and learns to select retrieve reliable information. Our module further enhances the performance of Top-k+MLP by 14.2\%.
This demonstrates that the proposed voting mechanism is effective for the tracking problem to maximize the positive effect of selecting candidates.

\noindent\textbf{Intermediate tracking data in memory.} 
To analyze the effectiveness of the memory in our framework, we train a model without storing any intermediate tracking information into the memory.
This model tracks the target object using only the initial frame.
\Tref{table:ablation} shows the AUC scores on the OTB-100 and UAV123 datasets.
With intermediate information in the memory, our framework obtains absolute gains of 4.8\% and 1.3\% on both datasets, respectively.

\begin{figure}[t]
\begin{center}

\includegraphics[width=0.9\linewidth]{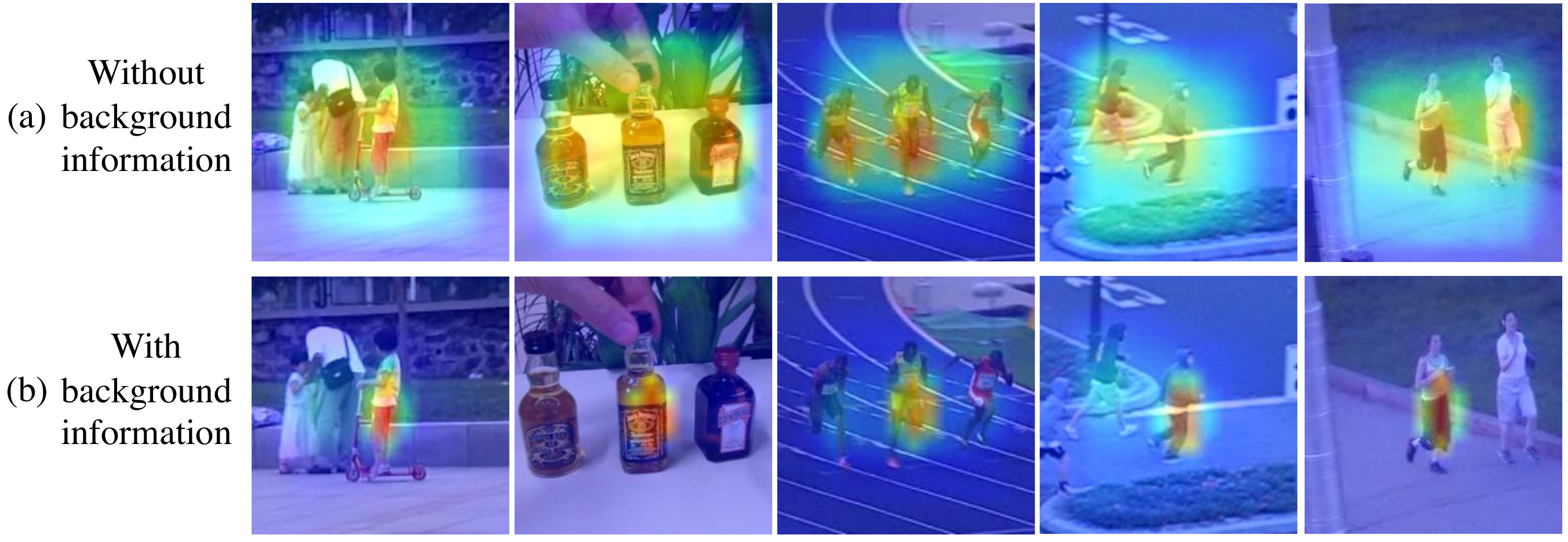}
\end{center}
\vspace{-0.6cm}
\caption{The effect of background information on the memory. We compare the score maps of our model with/without the background information.}
\vspace{-0.4cm}
\label{fig:bg_info}
\end{figure}
\noindent\textbf{Background information.}
We discuss the effect of the background information in the memory.
For the comparison, we run our model without the background information.
We zero-out the background pixels of the memory in this setting. 
The center prediction heatmaps are visualized in \fref{fig:bg_info}.
Without the background information, our tracker tends to be attracted by negative samples with the similar appearance.
This implicates that the background information in the memory is critical for our tracker to discriminate hard negatives better.

\subsection{Benchmark Results}
We compare our tracker with state-of-the-art methods on various tracking benchmarks: OTB-100~\cite{otb2015}, TrackingNet~\cite{trackingnet}, GOT-10k~\cite{got10k}, LaSOT~\cite{lasot}, and UAV123~\cite{mueller2016uav}.
We evaluate two different versions of our model, \textbf{\ours18} and \textbf{\ours50}, equipped with \texttt{resnet18} and \texttt{resnet50}, respectively.
We train and test the models using the same setting as explained in \Sref{subs:training} and \Sref{subs:inference}.
We use the same models for the evaluations on the OTB-100~\cite{otb2015}, LaSOT~\cite{lasot}, TrackingNet~\cite{trackingnet}, and UAV123~\cite{mueller2016uav} benchmarks.
For GOT-10k, we train a separate model using only the training split of GOT-10k according to its evaluation policy. 
For all other compared methods, we use officially released scores.

\begin{figure}[b]
\vspace{-0.6cm}
\begin{center}
\begin{subfigure}[t]{0.49\linewidth}
  \includegraphics[width=0.9\linewidth]{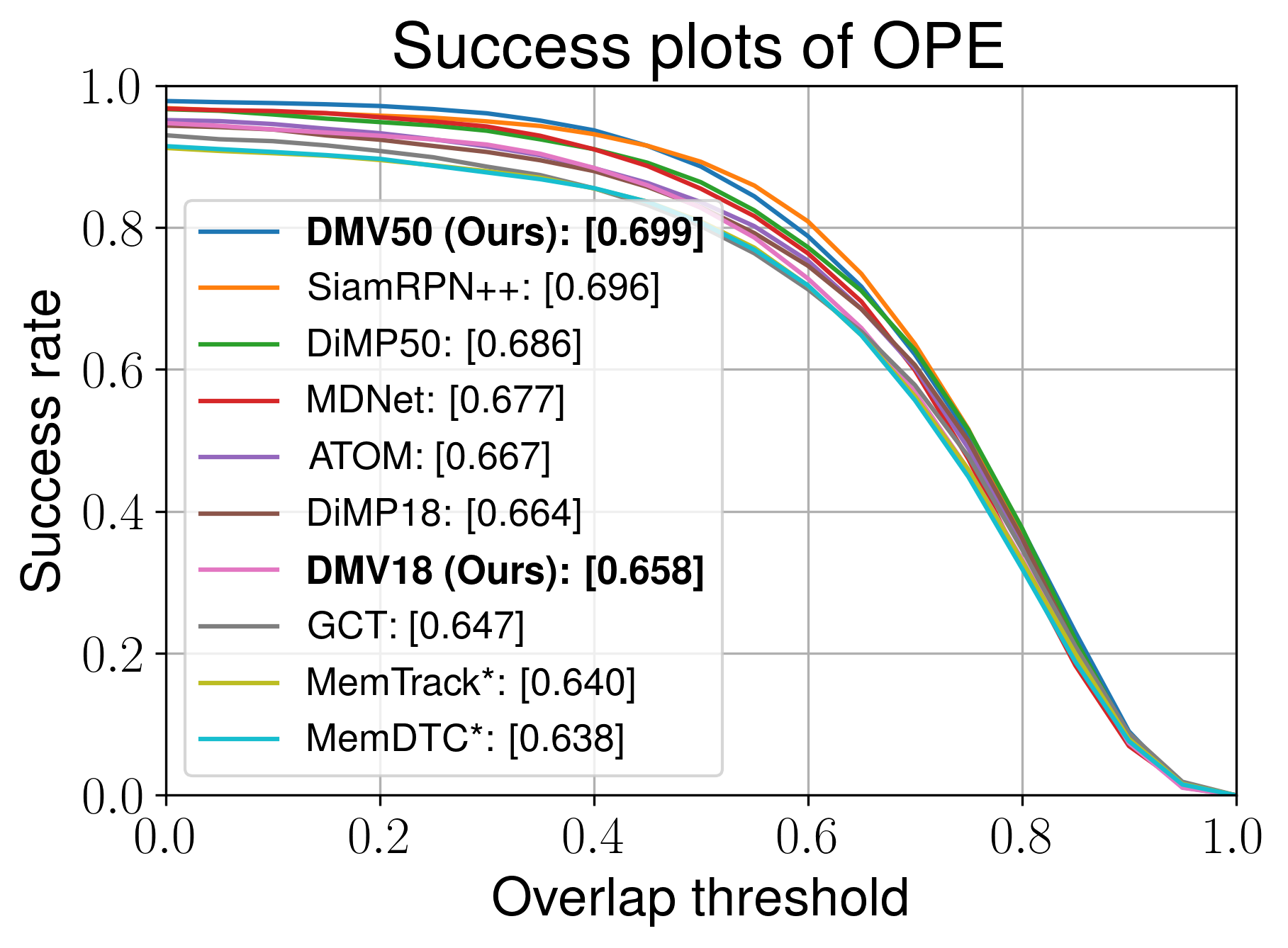}
\end{subfigure}
\begin{subfigure}[t]{0.49\linewidth}
  \includegraphics[width=0.9\linewidth]{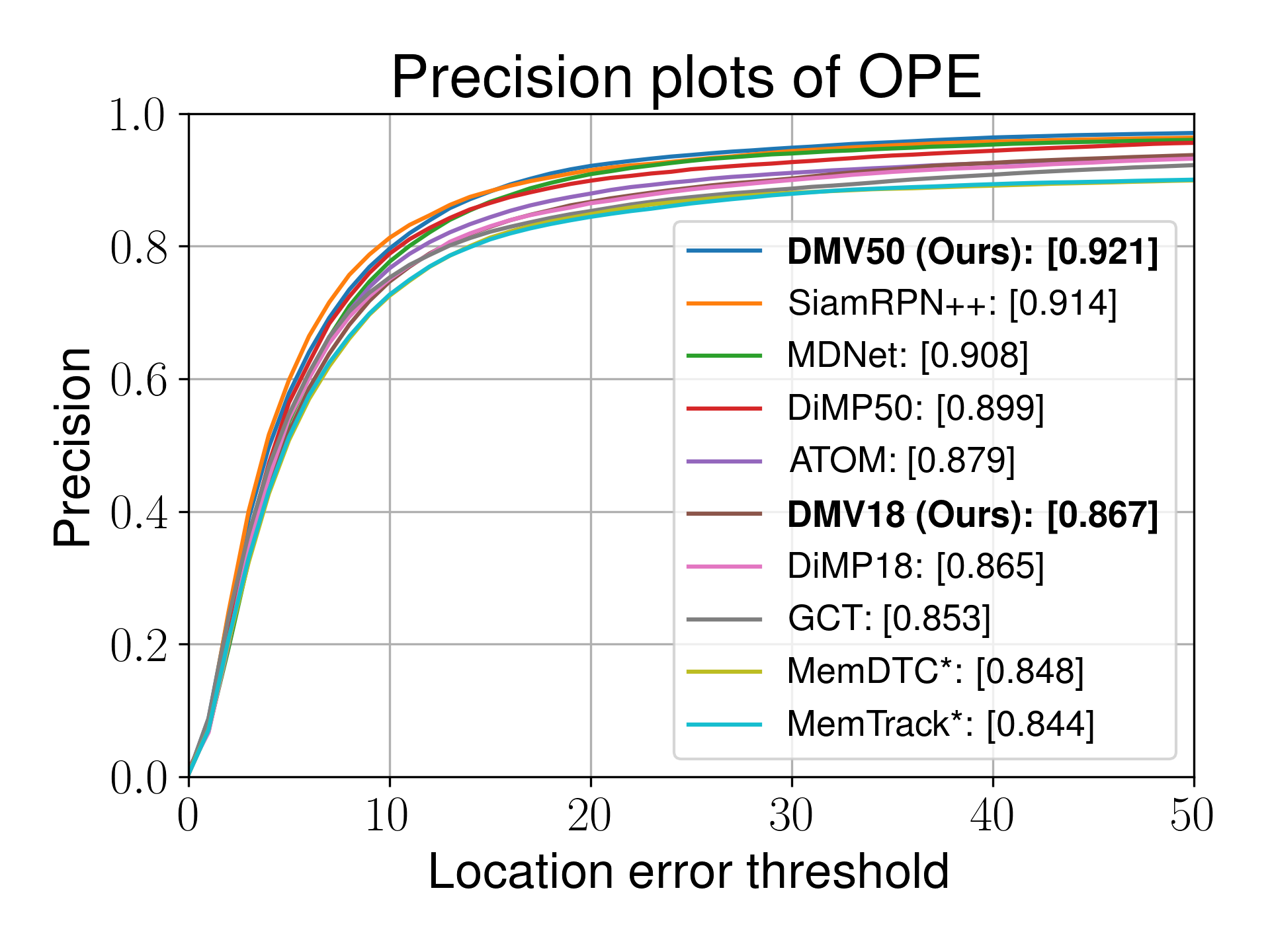}
\end{subfigure}
\end{center}
\vspace{-0.4cm}
\caption{Results on the OTB-100 dataset. 
}
\label{fig:otb}
\end{figure}

\Fref{fig:otb} shows the evaluation result on OTB-100 and \Tref{table:sota} summarizes the results on the other benchmarks.
In \Tref{table:sota}, the FPS of other methods are sourced from \cite{zhu2018dasiam}. We measure the FPS of our methods using NVIDIA TITAN X, which is the same GPU with one used in \cite{zhu2018dasiam}.
We colored the scores from top to third ranked methods in \textcolor{red}{\textbf{red}}, \textcolor{darkgreen}{\textbf{green}} and \textcolor{blue}{\textbf{blue}}, respectively.

\begin{table}[t]
\caption{Results on the various benchmarks.}

\setlength{\tabcolsep}{3pt}
\begin{adjustbox}{width=1\linewidth}

\begin{tabular}{l|ccc|ccc|cc|cc|c}

\toprule
& \multicolumn{3}{c|}{TrackingNet~\cite{trackingnet}} & \multicolumn{3}{c|}{GOT-10k~\cite{got10k}}  & \multicolumn{2}{c|}{LaSOT~\cite{lasot}} &
\multicolumn{2}{c|}{UAV123~\cite{mueller2016uav}}  &
\multirow{2}{*}{FPS} \\
\cmidrule{2-11}
& Succ. & Prec. & P$_{norm}$  & AO & SR$_{0.50}$ & SR$_{0.75}$ & Succ. & P$_{norm}$ & Succ. & Prec. & \\
\midrule

ECO~\cite{eco}       & 56.1 & 48.9 & 62.1 & 31.6 & 30.9 & 11.1 & 32.4 & 33.8  & 52.5 & 74.1 & 8\\ 
MDNet~\cite{mdnet}     & 61.4 & 55.5 & 71.0 & 29.9 & 30.3 & 9.9 & 39.7 & 46.0 &  - & - & 1\\
ATOM~\cite{danelljan2019atom}        & 70.3 & 64.8 & 77.1 & 55.6 & 63.4 & 40.2  & \textcolor{blue}{\textbf{51.5}}    & 57.6  & \textcolor{darkgreen}{\textbf{64.4}} &  \textcolor{darkgreen}{\textbf{83.2}} & 30 $^*$ \\ 
DiMP50~\cite{bhat2019dimp}     & \textcolor{blue}{\textbf{74.0}} & 68.7 & \textcolor{blue}{\textbf{80.1}}  & \textcolor{red}{\textbf{61.1}} & \textcolor{red}{\textbf{71.7}} & \textcolor{red}{\textbf{49.2}}&   \textcolor{red}{\textbf{56.9}}     & \textcolor{red}{\textbf{64.3}} & \textcolor{red}{\textbf{65.4}} &   \textcolor{red}{\textbf{83.7}}& 43 $^*$ \\

SiamFC~\cite{siamfc}     & 57.1 & 53.3 & 66.6  & 34.8 & 35.3 & 9.8 & 33.6 & 42.0 &  - & - & 86\\
SiamRPN~\cite{siamrpn}     & -     & -     & -   & 51.7 & 61.5 & 32.9 & - & -  & 52.7& 74.8 & 200  \\ 
DaSiamRPN~\cite{zhu2018dasiam}  & 63.8 & 59.1 & 73.3 & -     & -     & -    & 41.5 & 49.6 & 58.6  & 79.6 & 160\\
SiamRPN++~\cite{siamrpn++}  & 73.3 & \textcolor{blue}{\textbf{69.4}} & 80.0  & -     & -     & -    & 49.6 & 56.9  & 61.3 & \textcolor{blue}{\textbf{80.7}} & 35\\

\midrule
\textbf{\ours18 (Ours)} & \textcolor{darkgreen}{\textbf{75.2}} & \textcolor{darkgreen}{\textbf{70.0}} & \textcolor{darkgreen}{\textbf{80.4}}   & \textcolor{blue}{\textbf{56.9}} & \textcolor{blue}{\textbf{64.9}} & \textcolor{blue}{\textbf{45.7}} & \textcolor{darkgreen}{\textbf{53.7}} & \textcolor{darkgreen}{\textbf{60.8}} & 62.1 &  79.1 & 79\\ 

\textbf{\ours50 (Ours)} & \textcolor{red}{\textbf{75.6}} & \textcolor{red}{\textbf{71.4}} & \textcolor{red}{\textbf{81.6}} & \textcolor{darkgreen}{\textbf{60.1}} & \textcolor{darkgreen}{\textbf{69.5}} & \textcolor{red}{\textbf{49.2}} & 50.8 & \textcolor{blue}{\textbf{59.6}} & \textcolor{blue}{\textbf{62.8}}  &   80.4 & 46 \\ 
\bottomrule
\end{tabular}
\end{adjustbox}
\vspace{-0.3cm}
\begin{flushright}\footnotesize{$^*$ sourced from the original paper.}\end{flushright}
\label{table:sota}
\vspace{-0.6cm}
\end{table}

\vspace{2mm}
\noindent\textbf{OTB-100~\cite{otb2015}: }
The OTB-2015 dataset is the most popular benchmark dataset that most trackers are evaluated on. 
The dataset includes 100 sequences of various tracking scenarios.
We compare our method with various tracking algorithms, including memory-based trackers (MemTrack~\cite{yang2018memtrack}, MemDTC~\cite{yang2019memdtc}), online trackers (DiMP~\cite{bhat2019dimp}, MDNet~\cite{mdnet}, ATOM~\cite{danelljan2019atom}), and offline trackers (GCT~\cite{gao2019gct}, SiamRPN++~\cite{siamrpn++}).
\Fref{fig:otb} shows the results on the OTB-100 dataset.
While the previous memory-based trackers achieve relatively low performances, our \ourss, one of the memory-based trackers, achieves the best accuracy with 69.9\% and 92.1\% in the AUC scores of success and precision plots, respectively. 
Our method outperforms SiamRPN++~\cite{siamrpn++} as well as DiMP50~\cite{bhat2019dimp}, which are current state-of-the-arts offline and online trackers, respectively.

\noindent\textbf{TrackingNet~\cite{trackingnet}: }
TrackingNet is a large-scale benchmark dataset, built by re-purposing YouTube-BB~\cite{real2017youtube-bb} for the tracking task. The test split of the dataset consists of 511 sequences covering 20 object categories. 
We use the official evaluation server for the evaluation.
The server provides evaluation results with three measures: the area under curve score of success (Succ.), precision (Prec.), and the normalized precision (P$_{norm}$).
Our method outperforms all comparison methods.

\noindent\textbf{GOT-10k~\cite{got10k}: }
We use the test set of GOT-10k that consists of 180 short-term sequences.
We measure the average overlap (AO) and the success rate with two overlap thresholds (SR$_{0.50}$ and SR$_{0.75}$) using the evaluation server.
\ours50 achieves the best performance under the SR$_{0.75}$ metric and yields competitive results to the best method~\cite{bhat2019dimp} under the AO and SR$_{0.5}$ metrics.

\noindent\textbf{LaSOT~\cite{lasot}: }
LaSOT is one of the most challenging tracking benchmarks with very long sequences.
The test set of LaSOT consists of 280 videos and each video includes more than 2500 frames in average.
We report the results in terms of the AUC score of the success rate (Succ.) and the normalized distance precision (P$_{norm}$).
DiMP~\cite{bhat2019dimp} based on online learning shows the top accuracy in the benchmark.
SiamRPN++~\cite{siamrpn++} achieves competitive performance of 49.6\% and 56.9\% in terms of the AUC scores, without any online parameter tuning. 
\ours18 outperforms SiamRPN++ and yields competitive performance with DiMP50.

\noindent\textbf{UAV123~\cite{mueller2016uav}: }
UAV123 contains 123 sequences of altitude aerial videos.
Our tracker outperforms state-of-the-arts offline trackers including SiamRPN++ in terms of the AUC score of precision.

\noindent\textbf{Discussion: }
The experimental results show the strong potential of a memory-based approach in visual tracking. We achieved the best performance on OTB-100 and TrackingNet. Although our method showed relatively weak performance on UAV123, we conjecture that this is mainly because of the lack of training data. Our method relies on training data more than other methods (\eg DiMP50 and SiamRPN++) in order to learn a similarity metric for the memory retrieval. Current training data is dominated by the training set of TrackingNet, which is not so compatible with UAV123. We believe our method will benefit more from additional data than other methods.

\begin{figure}[t]
  \centering
  \includegraphics[width=1.0\linewidth]{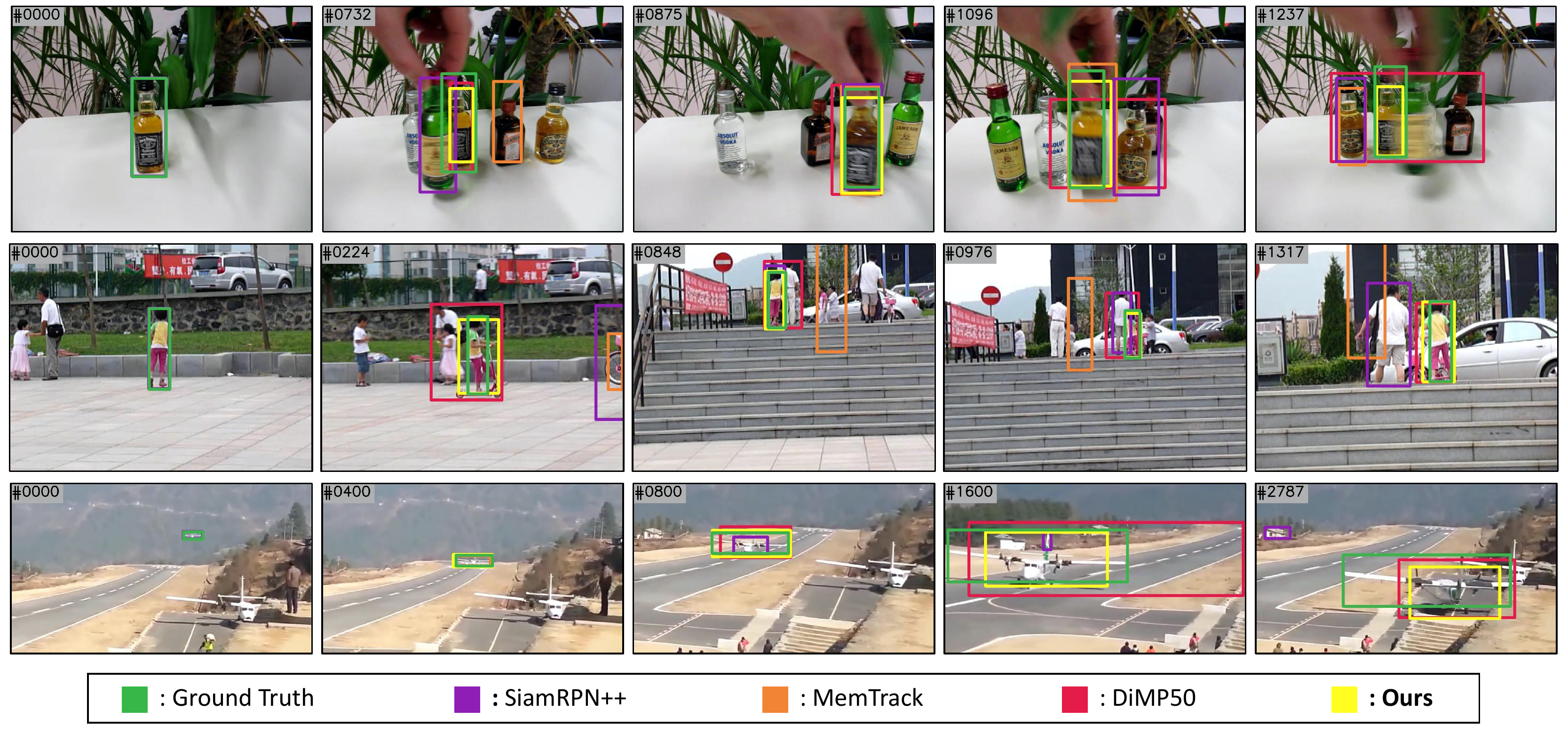}  
  
\vspace{-0.4cm}
\caption{Qualitative results of tracking methods.}
\label{fig:results_viz}
\vspace{-0.6cm}
\end{figure}

\Fref{fig:results_viz} shows some examples of the results from different tracking algorithms. Our method demonstrates robust tracking results in various challenging scenarios (\eg occlusion, appearance changes, and background clutter).

\section{Conclusion}
In this paper, we proposed a novel attention-based memory approach for visual object tracking.
The proposed memory-based tracker performs dense part matching followed by a novel voting mechanism to retrieve reliable information from the memory.
We demonstrated the strong potential of a memory-based approach in visual tracking and  achieved state-of-the-art results on various public benchmarks without online learning and explicit hard mining mining.

\vspace*{\stretch{1.0}}
\begin{center}
  \Large\textbf{Supplementary Materials: \\ DMV: Visual Object Tracking via Part-level Dense Memory and Voting-based Retrieval}\\
\end{center}
\vspace*{\stretch{2.0}}
   
This supplementary materials provide the details of our voting networks, further analysis and more results.
In \Sref{s:architecture}, the architectural details of the voting networks is described.
In \Sref{s:hyperparameters}, we perform ablation studies on important hyper-parameters.
In addition, \Sref{s:results} provides additional results on GOT-10k~\cite{got10k} and LaSOT~\cite{lasot}.

\begin{figure}[!b]
\begin{center}

\includegraphics[width=1\linewidth]{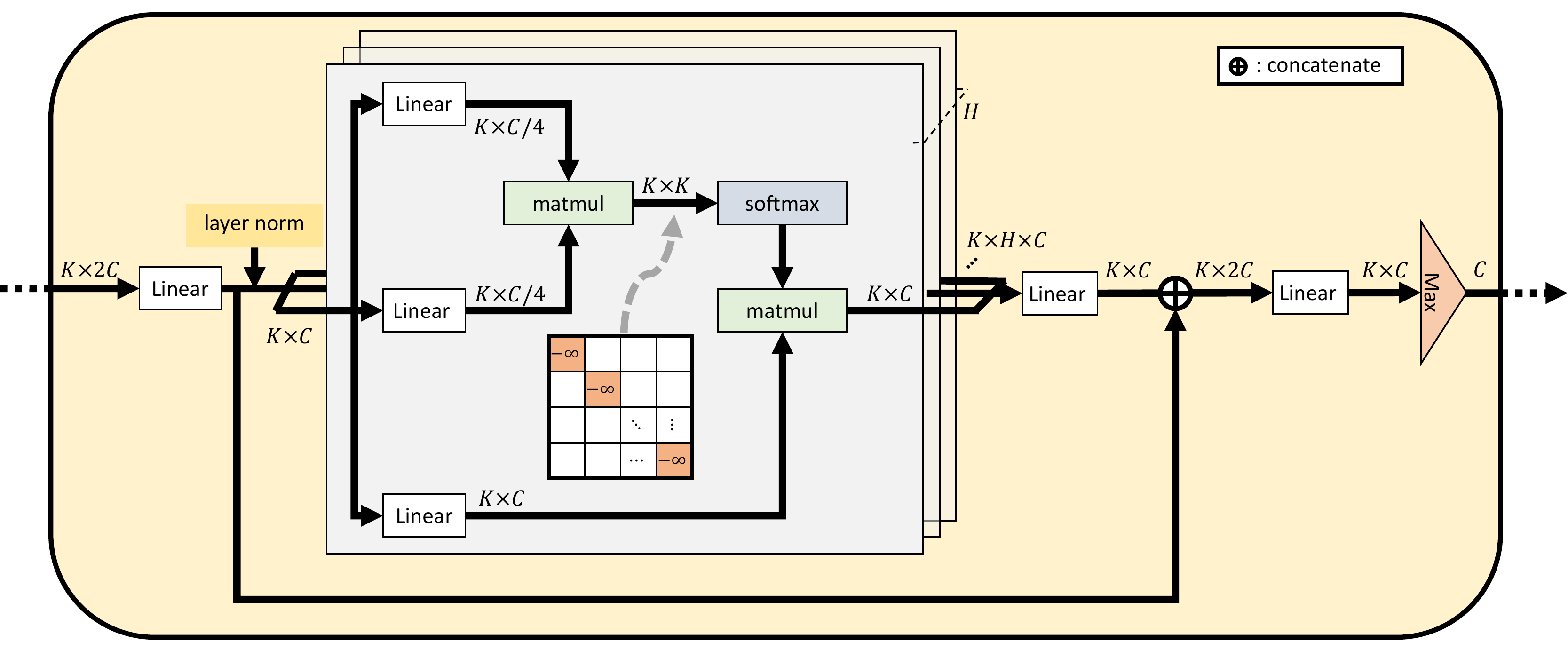}
\end{center}
\caption{The details of our voting networks.}
\label{fig:voting_networks}
\end{figure}

\section{Details of Voting Networks} \label{s:architecture}
The Transformer~\cite{vaswani2017attention} is used as the backbone for our voting networks. 
Here, we describe the detailed design of the transformer network and minor modifications we made on the original structure to meet our purpose.  
As shown in the detailed illustration (\Fref{fig:voting_networks}), we used a single layer of the multi-head transformer network as backbone. 
The number of heads ($H$) is set to 8. 
As the goal of our voting networks is to allow candidates to compare to each other, the self-attention is restricted by modifying values in the similarity matrix.
In specific, we reset the similarity scores toward themselves (diagonal elements) to minus infinity.

\section{Hyper-parameter Analysis} \label{s:hyperparameters}
In this section, we analyze the effects of hyper-parameters of our method.
We use \ours18 for the ablation studies. 

\begin{table*}[t]
\setlength{\tabcolsep}{5pt}
\centering
\caption{Results of AUC score on OTB-100 for each setting of $K$.}
\label{table:param_k}

\begin{adjustbox}{width=.7\textwidth}
\begin{tabular}{l|c|cccccc}
\toprule
\multicolumn{2}{c|}{OTB-100}& $K=1$ & $K=2$ & $K=4$ & $K=8$ & $K=16$ & $K=32$ \\
\midrule

\multicolumn{2}{c|}{AUC score} & 63.6 & 62.6 & 65.8 & 66.6 & 68.7 & 69.2 \\
\multicolumn{2}{c|}{FPS} & 87 & 82 & 79 & 72 & 61 & 44 \\
\bottomrule
\end{tabular}
\end{adjustbox}
\end{table*}

\noindent\textbf{The number of candidates. }
The number of candidates $K$ in our voting mechanism is one of most important hyper-parameters.
To investigate the effect of $K$, we train and test our model with different $K$'s.
\Tref{table:param_k} shows the resulting AUC scores and the running speed on OTB-100 for each $K$.
It is clearly visible that there is a trade-off between the performance and the speed.
While using a large $K$ leads a steady improvement in the accuracy, FPS is also decreased accordingly.
We set $K=4$ for all the experiments in the main paper as it shows the best trade-off.

\begin{table*}[t]
\setlength{\tabcolsep}{5pt}
\centering
\caption{Ablation study on the memory capacity and the frame interval.}
\label{table:cap_rate}

\begin{adjustbox}{width=0.7\textwidth}
\begin{tabular}{c}

\includegraphics[width=1\linewidth]{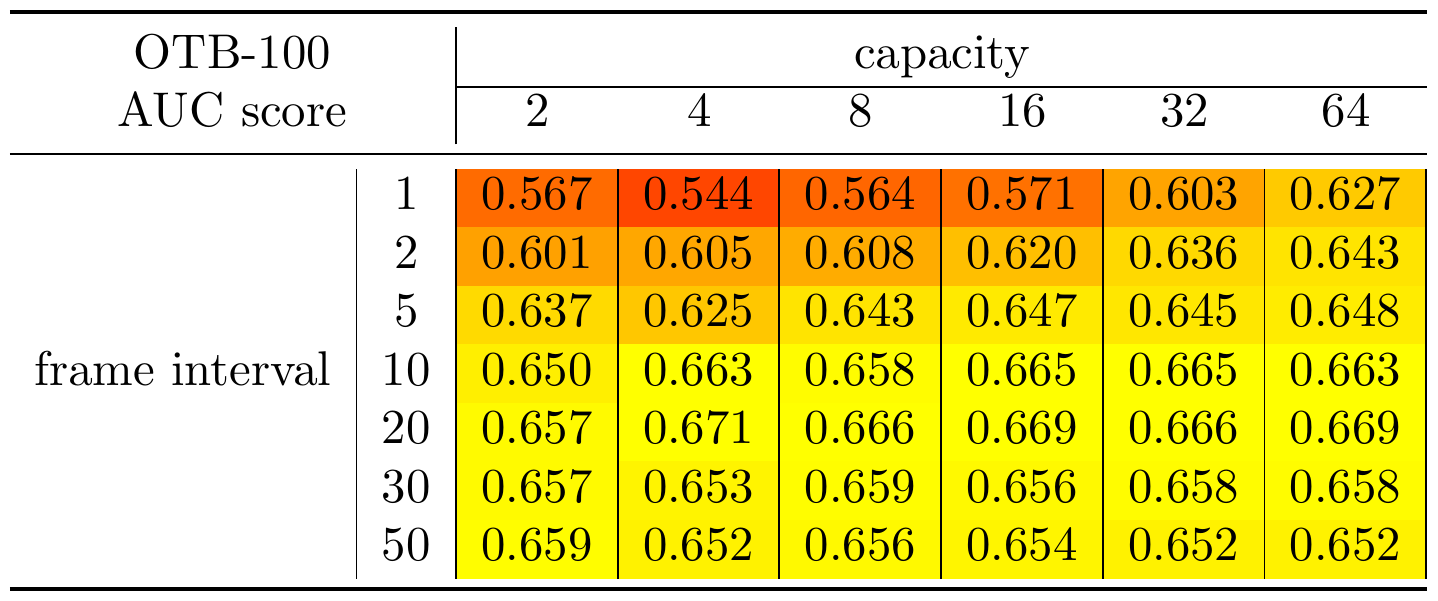}

\end{tabular}
\end{adjustbox}
\end{table*}

\noindent\textbf{Capacity vs. frame interval. }
The capacity of the memory and the time interval between memory frames (the frequency of saving new memory) are key hyper-parameters for the testing.
These parameters define the memory management rule during the inference.
\Tref{table:cap_rate} shows the changes of the AUC score on OTB-100 according to different memory capacities and frame intervals.
When either the memory capacity or the frame interval is too low, the performance drops significantly (e.g. table cells in reddish color).
Our method steadily shows acceptable performance when both of the hyper-parameters are in the range of proper values (e.g. capacity $>$ 8 $\cap$ interval $>$ 10). 
In the main paper, we set the frame interval as 30 and the memory capacity as 32 in all the experiments.

\section{More Results} \label{s:results}
\begin{figure}[t]
\begin{center}

\includegraphics[width=.8\linewidth]{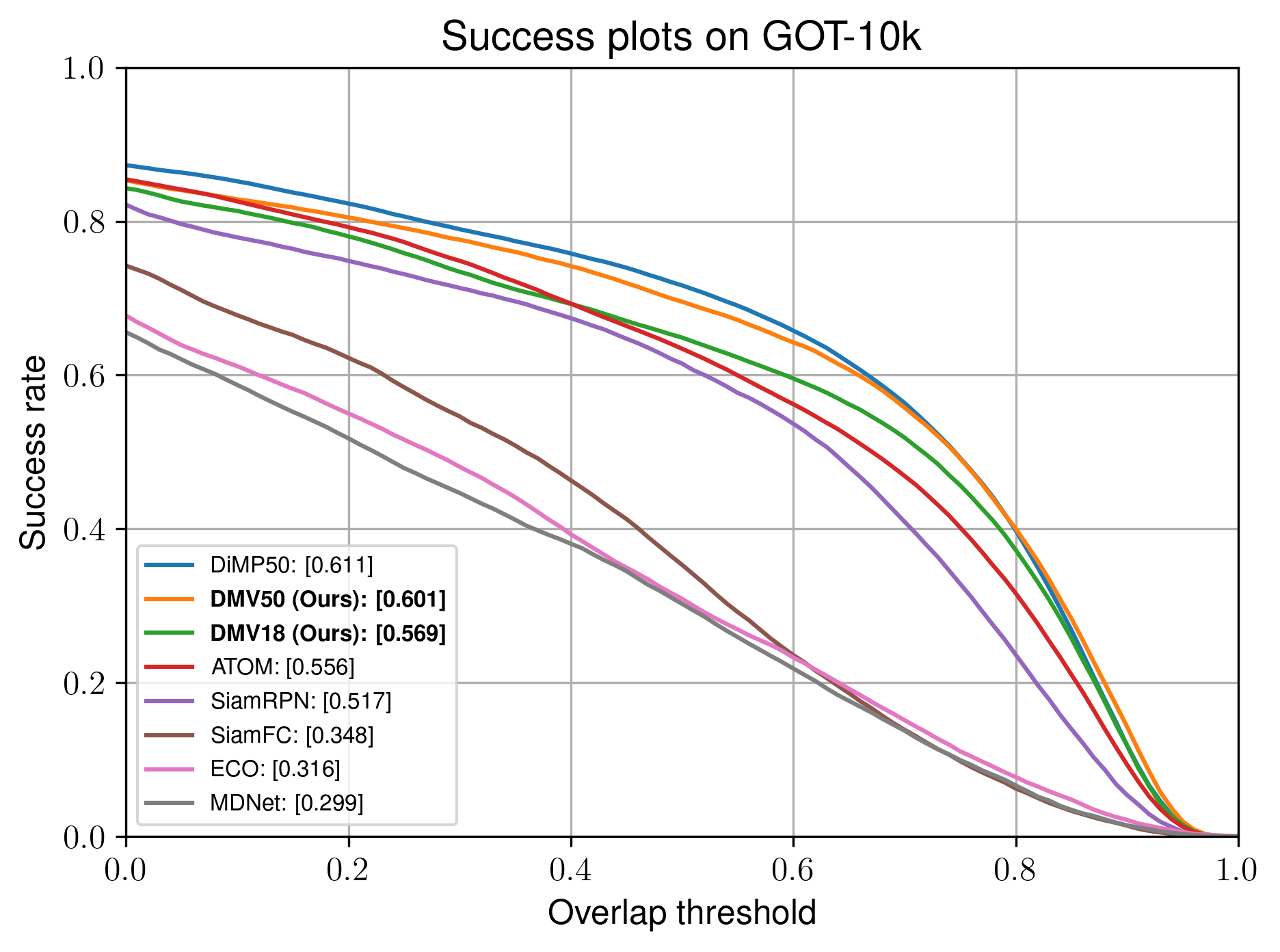}
\end{center}
\caption{Success plots on GOT-10k test set.}
\label{fig:got_results}
\end{figure}

\begin{figure}[t]
\begin{center}

\includegraphics[width=1\linewidth]{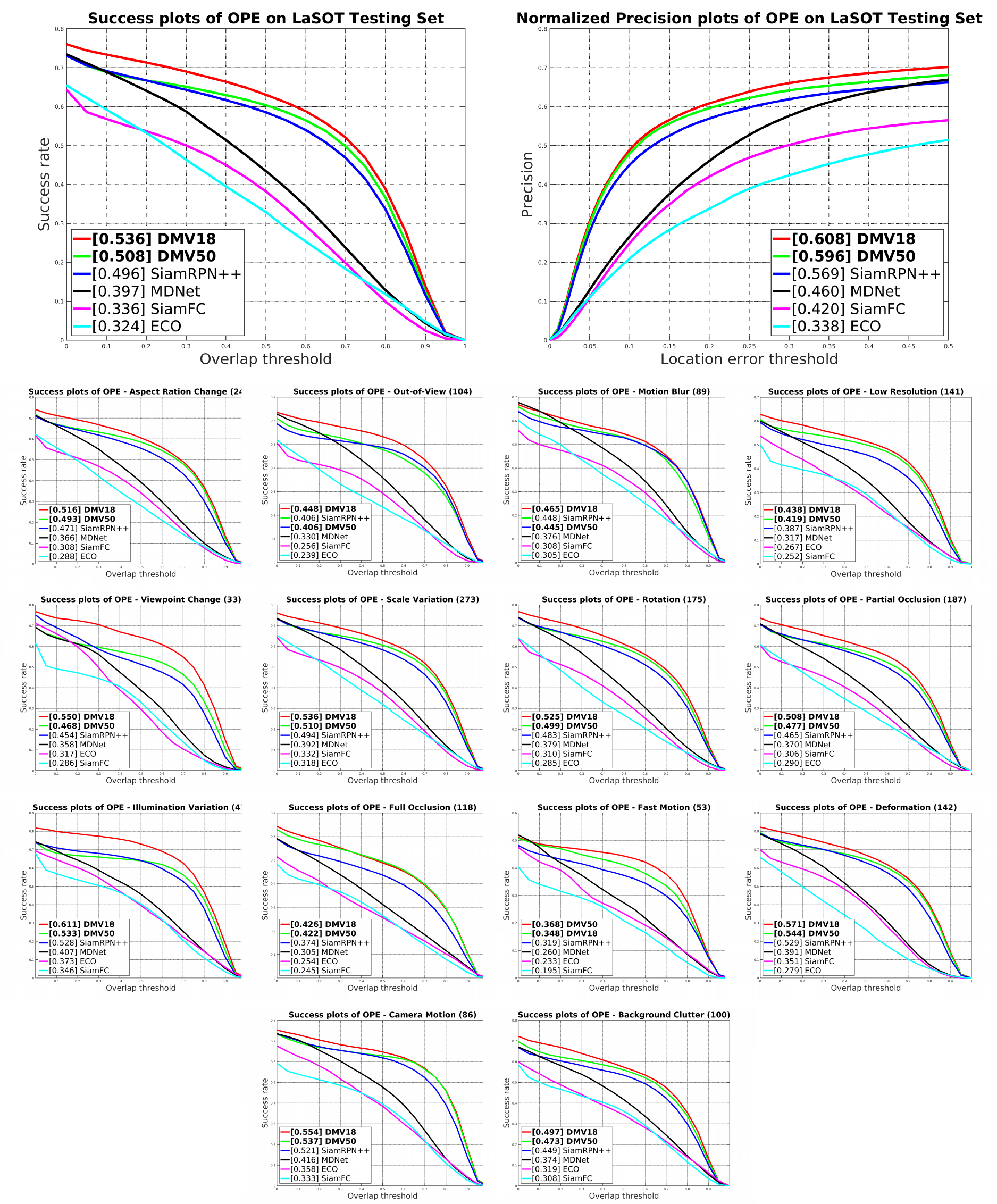}
\end{center}
\caption{Results on LaSOT dataset including attribute analysis.}
\label{fig:lasot_results}
\end{figure}

In this section, we provide additional result plots on GOT-10k~\cite{got10k} and LaSOT~\cite{lasot}. 
\Fref{fig:got_results} shows the success plots on GOT-10k~\cite{got10k} test set.
The success plots and normalized precision plots of offline learning methods on the LaSOT~\cite{lasot} dataset are also provided in \Fref{fig:lasot_results}.
Additionally, we report 14 different attributes for each video sequence on the LaSOT dataset: aspect ratio change, out-of-view, motion blur, low resolution, viewpoint change, scale variation, rotation, partial occlusion, illumination variation, full occlusion, fast motion, deformation, camera motion, background clutter.


\clearpage

\clearpage
%
%
\bibliographystyle{utils/splncs04}
\bibliography{egbib}

\end{document}